\documentclass[letterpaper, 10 pt, conference]{ieeeconf}  % Comment this line out if you need a4paper

\IEEEoverridecommandlockouts                              % This command is only needed if 
\usepackage[english]{babel}

\usepackage[letterpaper,top=2cm,bottom=2cm,left=3cm,right=3cm,marginparwidth=1.75cm]{geometry}

\usepackage{amsmath}
\usepackage{graphicx}
\usepackage[colorlinks=true, allcolors=blue]{hyperref}
\usepackage{listings}
\usepackage{cleveref}

\title{\LARGE \bf
Advancing The Robotics Software Development Experience: Bridging Julia's Performance and Python's Ecosystem
}

\author{Gustavo Nunes Goretkin, Joseph Carpinelli and Andy Park} % <-this % stops a space
\begin{document}

\maketitle
\thispagestyle{empty}
\pagestyle{empty}

%%%%%%%%%%%%%%%%%%%%%%%%%%%%%%%%%%%%%%%%%%%%%%%%%%%%%%%%%%%%%%%%%%%%%%%%%%%%%%%%
\begin{abstract}

Robotics programming typically involves a trade-off between the ease of use offered by Python and the run-time performance of C++. While multi-language architectures address this trade-off by coupling Python's ergonomics with C++'s speed, they introduce complexity at the language interface. This paper proposes using Julia for performance-critical tasks within Python ROS 2 applications, providing an elegant solution that streamlines the development process without disrupting the existing Python workflow.

\end{abstract}

%%%%%%%%%%%%%%%%%%%%%%%%%%%%%%%%%%%%%%%%%%%%%%%%%%%%%%%%%%%%%%%%%%%%%%%%%%%%%%%%
\section{Introduction}

Python and C++ serve as the foundational languages in robotics programming, each providing unique advantages—Python for its ergonomics and C++ for computational efficiency. This duality has led to prevalent multi-language architectures where C++ handles core computations and Python is used for higher-level integration via wrappers. The Python wrappers are then used for integration (or "glue") code, as well as interactive exploration and prototyping. This general approach of integrating a fast compiled language and an ergonomic scripting language has been the bedrock of technical computing for decades, largely popularized by MATLAB providing wrappers to linear algebra routines written in FORTRAN. Performance in this paradigm is achieved by "vectorizing the code", ensures that the performance-critical loops are in the fast language (in computational "kernels"), rather than in the slow host language. Many robotics libraries follow this paradigm, including Drake~\cite{drake}, GTSAM~\cite{dellaert2012factor}, OpenCV~\cite{opencv_library}, Open3D~\cite{Zhou2018}.

The Robot Operating System (ROS)~\cite{ros2} exemplifies another approach, facilitating Python and C++ interoperation by relying on an interface definition language (IDL) to generate code in both languages. The distributed system approach offers other advantages, such as reducing coupling and promoting scalability, and it aligns with the microservices architecture popular in software engineering more broadly. Generating C/C++ code that performs substantial computation (as opposed to just serving as an interface) is yet another approach taken by some libraries and systems \cite{symforce,open2020,mattingley2012cvxgen}.

These traditional Python-C++ architectures, while robust, come with their own set of challenges, particularly in terms of development experience, system complexity, and performance overhead. This context sets the stage for considering alternative approaches that might streamline development without sacrificing run-time performance. Julia, designed for both high performance and ease of use~\cite{julia_fresh}, emerges as a promising alternative. 
Despite its potential, Julia's current ecosystem limitations, such as the absence of a ROS 2 client library, pose challenges for its adoption in robotics. We propose an architecture that leverages the mature ecosystem of Python alongside Julia, reducing the reliance on C++. This approach maintains a two-language paradigm but shifts the focus from complementary technical benefits to leveraging social and ecosystem advantages. Moreover, the compatibility between Julia and Python, facilitated by features like interactivity and the characteristics of dynamic typing, minimizes the development mismatch. We further explore Julia's application in robotics, illustrating its integration with Python.

\subsection{Python wrappers of C++ code}

Python bindings to C++ code, primarily facilitated by pybind11~\cite{pybind11}, bridge the two languages but introduce complexity in type conversions and memory management. Alternatively, cppyy~\cite{lavrijsen2016high,kundu2023efficient} integrates directly with the C++ compiler for improved performance through JIT compilation and support for generic programming with C++ templates. Both methods aim to combine Python's ease with C++'s efficiency, albeit with challenges in debugging and optimization.

\subsection{Ergonomics of a programming system}

Programming effectiveness hinges not just on language syntax and semantics but on the entire \emph{system}, which includes the development experience, library ecosystems, and package management~\cite{evaluating-systems,Jakubovic2023TechnicalDimensionsOfProgrammingSystems}.
These broader aspects play a crucial role in determining how efficiently and comfortably developers can work within a programming environment.

C++ is often favored for its static type safety and the precise control it offers over memory allocation, features that are critical for high-performance computing and systems-level programming. However, Python is widely regarded as more user-friendly, making it the preferred choice for prototyping, data analysis, visualization, and exploratory programming. A significant factor behind Python's ease of use is its read-evaluate-print-loop (REPL), which is commonly accessed through notebook interfaces like Jupyter~\cite{kluyver2016jupyter}. This interactive environment facilitates rapid testing and iteration, a key advantage in exploratory programming.

Despite ongoing efforts to enhance Python's efficiency through various implementations (e.g. Numba, Pyston, Cinder, pypy, pyston, pydjion, Nuitka, Shed Skin, etc.)~\cite{spiegelberg2021tuplex,lam2015numba,python_cinder,python_shedskin,python_nukita,python_pyjion}, and to make C++ more interactive \cite{vasilev2012cling,antcheva2009root}, the reality is that achieving a blend of interactivity and performance requires intentional and often divergent design considerations.
A particular challenge with Python lies in its reliance on the CPython C API for most of its ecosystem. This dependency has historically limited the language's ability to achieve high runtime performance. The CPython C API is deeply integrated into the Python ecosystem, making any changes to it potentially disruptive. As a result, while there have been significant efforts to improve Python's performance, the fundamental architecture of its most common implementation presents a bottleneck

\section{The design of Julia}
From its inception, the design of Julia has focused on integration between an interactive environment and a compiler. The design has continued to improve, particularly in relation to caching compilation and interactively invalidating it. At the heart of Julia's performance and ergonomic advantages are two key features: staged programming, which leverages compilation built on LLVM, and its pervasive use of multiple dispatch. Multiple dispatch is distinct from function overloading seen in languages like C++ and Java in that Julia does not have a semantic distinction between compile-time and run-time types. This strategy not only enhances runtime efficiency but also significantly benefits developer ergonomics by promoting extensive code reuse and enabling serendipitous composition, though the correctness of the composition must still be verified~\cite{julia_fresh}.

\section{Julia-Python interoperation}

The Julia package ecosystem hosts several projects which can help developers write cross-language software. The \texttt{JuliaPy} organization facilitates cross-language development between Julia and Python, at first offering \texttt{PyCall.jl} and \texttt{pyjulia}~\cite{GitHubJuliaPyOrg}. More recently, \texttt{JuliaPy} expanded its toolkit with \texttt{PythonCall.jl} and \texttt{juliacall}, enhancing the integration while simplifying dependency management~\cite{pythoncall-docs}. \texttt{juliacall}, available via \href{https://pypi.org}{PyPI}, enables direct Julia calls within Python without custom bindings, leveraging \texttt{import juliacall} for seamless language interoperation. This approach streamlines the use of Julia's capabilities within Python environments, facilitating the incorporation of Julia's computational efficiency into Python-based applications. Several popular Python projects have utilized Julia backends through \texttt{juliacall}
\cite{DifferentialEquations.jl-2017,cranmerInterpretableMachineLearning2023}.

\section{Case Study: {\texttt{FlexIK.jl}} and {\texttt{flex\_ik\_py}}}
\label{sec:case-study}

In this case study, we delve into the comparative analysis of two implementations of a numerical Inverse Kinematics (IK) solver tailored for the Spot robot with the Spot Arm: {\texttt{FlexIK.jl}}, created using Julia, and {\texttt{flex\_ik\_py}}, developed with Python. The core of these implementations is the a prioritized stack of tasks framework~\cite{Fiore2023IkWithArbitraryConstraints}, adeptly managing a whole-body IK challenge given 6 Degrees of Freedom (DOF) user input, controlling 18 joints of the robot. This dual-task strategy ensures tracking of the end-effector pose while minimizing torso movement, facilitating natural adjustments when targets are beyond arm's reach alone, thus broadening the operational capabilities of the robot.

\vspace{-0.1 in}
\begin{figure}[ht]
\centering
\includegraphics[width=0.47\textwidth]{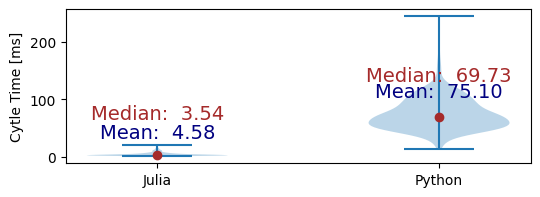}
\caption{Comparison of computation times between the FlexIK solver implementations in Julia and Python.}
\label{fig:flexik_comparison}
\end{figure}
\vspace{-0.1 in}

Figure \ref{fig:flexik_comparison} reveals that the median computation time for {\texttt{FlexIK.jl}} is 3.49 ms, significantly lower than the 66.76 ms for the Python counterpart, highlighting Julia's over 20x improvement in performance. This notable difference not only illustrates Julia's efficiency and stability for real-time robotics but also its impact on robot behavior, leading to smoother and more responsive actions. It is noteworthy that the Julia implementation's performance was achieved without specific efforts to optimize the code. The {\texttt{FlexIK.jl}} versus {\texttt{flex\_ik\_py}} case study underscores Julia's potential to advance robotics programming, advocating for deeper exploration into its capabilities for complex system development. 

While \texttt{FlexIK.jl} executes substantially faster than \texttt{flex\_ik\_py}, Python bindings to Julia do impose their own set of limitations. At the time of writing, \texttt{juliacall} may only run in a single Python thread. In addition, while not unique to Julia, garbage collection pauses the \texttt{FlexIK.jl} implementation, and therefore negatively impacts performance. Further improvements to the Julia implementation, including reductions in memory allocations, would further improve performance.

\section{Conclusion}

The integration of Julia into a Python-based robotics system presents a pragmatic approach to improving computational efficiency while maintaining development workflow. Our case study with {\texttt{FlexIK.jl}} and {\texttt{flex\_ik\_py}} demonstrates Julia's significant performance advantages, with computation times improved by an order of magnitude. These results suggest that Julia can serve as a viable alternative to the traditional Python-C++ paradigm, reducing development complexity without sacrificing speed. Future work should focus on expanding Julia's ecosystem, particularly in ROS 2 environments, to fully leverage its potential in robotics programming.

%%%%%%%%%%%%%%%%%%%%%%%%%%%%%%%%%%%%%%%%%%%%%%%%%%%%%%%%%%%%%%%%%%%%%%%%%%%%%%%%

\bibliographystyle{IEEEtran}
\bibliography{ref}

\begin{thebibliography}{10}
\providecommand{\url}[1]{#1}
\csname url@samestyle\endcsname
\providecommand{\newblock}{\relax}
\providecommand{\bibinfo}[2]{#2}
\providecommand{\BIBentrySTDinterwordspacing}{\spaceskip=0pt\relax}
\providecommand{\BIBentryALTinterwordstretchfactor}{4}
\providecommand{\BIBentryALTinterwordspacing}{\spaceskip=\fontdimen2\font plus
\BIBentryALTinterwordstretchfactor\fontdimen3\font minus
  \fontdimen4\font\relax}
\providecommand{\BIBforeignlanguage}[2]{{%
\expandafter\ifx\csname l@#1\endcsname\relax
\typeout{** WARNING: IEEEtran.bst: No hyphenation pattern has been}%
\typeout{** loaded for the language `#1'. Using the pattern for}%
\typeout{** the default language instead.}%
\else
\language=\csname l@#1\endcsname
\fi
#2}}
\providecommand{\BIBdecl}{\relax}
\BIBdecl

\bibitem{drake}
\BIBentryALTinterwordspacing
R.~Tedrake and the Drake Development~Team, ``Drake: Model-based design and
  verification for robotics,'' 2019. [Online]. Available:
  \url{https://drake.mit.edu}
\BIBentrySTDinterwordspacing

\bibitem{dellaert2012factor}
F.~Dellaert, ``Factor graphs and {GTSAM}: {A} hands-on introduction,''
  \emph{Georgia Institute of Technology, Tech. Rep}, vol.~2, p.~4, 2012.

\bibitem{opencv_library}
G.~Bradski, ``{The OpenCV Library},'' \emph{Dr. Dobb's Journal of Software
  Tools}, 2000.

\bibitem{Zhou2018}
Q.-Y. Zhou, J.~Park, and V.~Koltun, ``{Open3D}: {A} modern library for {3D}
  data processing,'' \emph{arXiv:1801.09847}, 2018.

\bibitem{ros2}
\BIBentryALTinterwordspacing
S.~Macenski, T.~Foote, B.~Gerkey, C.~Lalancette, and W.~Woodall, ``Robot
  {O}perating {S}ystem 2: {D}esign, architecture, and uses in the wild,''
  \emph{Science Robotics}, vol.~7, no.~66, p. eabm6074, 2022. [Online].
  Available: \url{https://www.science.org/doi/abs/10.1126/scirobotics.abm6074}
\BIBentrySTDinterwordspacing

\bibitem{symforce}
\BIBentryALTinterwordspacing
H.~Martiros, A.~Miller, N.~Bucki, B.~Solliday, R.~Kennedy, J.~Zhu, T.~Dang,
  D.~Pattison, H.~Zheng, T.~Tomic, P.~Henry, J.~VanderMey, G.~Cross, A.~Sun,
  S.~Wang, and K.~Holtz, ``Symforce: Symbolic computation and code generation
  for robotics,'' in \emph{Proceedings of Robotics: Science and Systems}, 2022.
  [Online]. Available: \url{https://github.com/symforce-org/symforce}
\BIBentrySTDinterwordspacing

\bibitem{open2020}
P.~Sopasakis, E.~Fresk, and P.~Patrinos, ``{OpEn}: Code generation for embedded
  nonconvex optimization,'' in \emph{IFAC World Congress}, Berlin, 2020.

\bibitem{mattingley2012cvxgen}
J.~Mattingley and S.~Boyd, ``{CVXGEN}: A code generator for embedded convex
  optimization,'' \emph{Optimization and Engineering}, vol.~13, pp. 1--27,
  2012.

\bibitem{julia_fresh}
\BIBentryALTinterwordspacing
J.~Bezanson, A.~Edelman, S.~Karpinski, and V.~B. Shah, ``Julia: {A} fresh
  approach to numerical computing,'' \emph{SIAM Review}, vol.~59, no.~1, pp.
  65--98, 2017. [Online]. Available: \url{https://doi.org/10.1137/141000671}
\BIBentrySTDinterwordspacing

\bibitem{pybind11}
W.~Jakob, J.~Rhinelander, and D.~Moldovan, ``{pybind11} -- seamless operability
  between {C++11} and {P}ython,'' 2017, https://github.com/pybind/pybind11.

\bibitem{lavrijsen2016high}
W.~T. Lavrijsen and A.~Dutta, ``High-performance {P}ython-{C}++ bindings with
  {PyPy} and {C}ling,'' in \emph{2016 6th Workshop on Python for
  High-Performance and Scientific Computing (PyHPC)}.\hskip 1em plus 0.5em
  minus 0.4em\relax IEEE, 2016, pp. 27--35.

\bibitem{kundu2023efficient}
B.~Kundu, V.~Vassilev, and W.~Lavrijsen, ``Efficient and accurate automatic
  {P}ython bindings with cppyy \& {C}ling,'' \emph{arXiv preprint
  arXiv:2304.02712}, 2023.

\bibitem{evaluating-systems}
J.~Edwards, S.~Kell, T.~Petricek, and L.~Church, ``Evaluating programming
  systems design,'' in \emph{Proceedings of 30th Annual Workshop of Psychology
  of Programming Interest Group}, ser. PPIG 2019, 2019.

\bibitem{Jakubovic2023TechnicalDimensionsOfProgrammingSystems}
\BIBentryALTinterwordspacing
J.~Jakubovic, J.~Edwards, and T.~Petricek, ``Technical dimensions of
  programming systems,'' \emph{The Art, Science, and Engineering of
  Programming}, vol.~7, no.~3, 2023. [Online]. Available:
  \url{https://programming-journal.org/2023/7/13/}
\BIBentrySTDinterwordspacing

\bibitem{kluyver2016jupyter}
T.~Kluyver, B.~Ragan-Kelley, F.~P{\'e}rez, B.~E. Granger, M.~Bussonnier,
  J.~Frederic, K.~Kelley, J.~B. Hamrick, J.~Grout, S.~Corlay \emph{et~al.},
  ``Jupyter notebooks-a publishing format for reproducible computational
  workflows.'' \emph{Elpub}, vol. 2016, pp. 87--90, 2016.

\bibitem{spiegelberg2021tuplex}
L.~Spiegelberg, R.~Yesantharao, M.~Schwarzkopf, and T.~Kraska, ``Tuplex: Data
  science in python at native code speed,'' in \emph{Proceedings of the 2021
  International Conference on Management of Data}, 2021, pp. 1718--1731.

\bibitem{lam2015numba}
S.~K. Lam, A.~Pitrou, and S.~Seibert, ``Numba: A {LLVM}-based python {JIT}
  compiler,'' in \emph{Proceedings of the Second Workshop on the LLVM Compiler
  Infrastructure in HPC}, 2015, pp. 1--6.

\bibitem{python_cinder}
\BIBentryALTinterwordspacing
Meta, ``Cinder is {M}eta's internal performance-oriented production version of
  {CP}ython.'' 2021. [Online]. Available:
  \url{https://github.com/facebookincubator/cinder}
\BIBentrySTDinterwordspacing

\bibitem{python_shedskin}
\BIBentryALTinterwordspacing
``Shed skin is a restricted-{P}ython-to-{C}++ compiler.'' 2010. [Online].
  Available: \url{https://github.com/shedskin/shedskin}
\BIBentrySTDinterwordspacing

\bibitem{python_nukita}
\BIBentryALTinterwordspacing
``Nuitka is a {P}ython compiler written in {P}ython.'' 2011. [Online].
  Available: \url{https://github.com/Nuitka/Nuitka}
\BIBentrySTDinterwordspacing

\bibitem{python_pyjion}
\BIBentryALTinterwordspacing
``Pyjion - {A} {JIT} for {P}ython based upon {C}ore{CLR},'' 2015. [Online].
  Available: \url{https://github.com/tonybaloney/pyjion}
\BIBentrySTDinterwordspacing

\bibitem{vasilev2012cling}
V.~Vasilev, P.~Canal, A.~Naumann, and P.~Russo, ``Cling--the new interactive
  interpreter for {ROOT} 6,'' in \emph{Journal of Physics: Conference Series},
  vol. 396, no.~5.\hskip 1em plus 0.5em minus 0.4em\relax IOP Publishing, 2012,
  p. 052071.

\bibitem{antcheva2009root}
I.~Antcheva, M.~Ballintijn, B.~Bellenot, M.~Biskup, R.~Brun, N.~Buncic,
  P.~Canal, D.~Casadei, O.~Couet, V.~Fine \emph{et~al.}, ``Root—{A} {C++}
  framework for petabyte data storage, statistical analysis and
  visualization,'' \emph{Computer Physics Communications}, vol. 180, no.~12,
  pp. 2499--2512, 2009.

\bibitem{GitHubJuliaPyOrg}
{JuliaPy}, ``Software that connects the {J}ulia and {P}ython languages.''
  \url{https://github.com/JuliaPy}.

\bibitem{pythoncall-docs}
\BIBentryALTinterwordspacing
JuliaPy, ``{PythonCall \& JuliaCall},'' package documentation for PythonCall
  and JuliaCall. [Online]. Available:
  \url{https://juliapy.github.io/PythonCall.jl/v0.9.15/}
\BIBentrySTDinterwordspacing

\bibitem{DifferentialEquations.jl-2017}
\BIBentryALTinterwordspacing
C.~Rackauckas and Q.~Nie, ``Differential{E}quations.jl – {A} performant and
  feature-rich ecosystem for solving differential equations in julia,''
  \emph{The Journal of Open Research Software}, vol.~5, no.~1, 2017, exported
  from https://app.dimensions.ai on 2019/05/05. [Online]. Available:
  \url{https://app.dimensions.ai/details/publication/pub.1085583166}
\BIBentrySTDinterwordspacing

\bibitem{cranmerInterpretableMachineLearning2023}
\BIBentryALTinterwordspacing
M.~Cranmer, ``Interpretable {Machine} {Learning} for {Science} with {PySR} and
  {SymbolicRegression}.jl,'' May 2023, arXiv:2305.01582 [astro-ph,
  physics:physics]. [Online]. Available: \url{http://arxiv.org/abs/2305.01582}
\BIBentrySTDinterwordspacing

\bibitem{Fiore2023IkWithArbitraryConstraints}
M.~D. Fiore, G.~Meli, A.~Ziese, B.~Siciliano, and C.~Natale, ``A general
  framework for hierarchical redundancy resolution under arbitrary
  constraints,'' \emph{IEEE Transactions on Robotics}, 2023.

\end{thebibliography}

\end{document}